\documentclass[fleqn]{article}

\usepackage[textwidth=17.6cm]{geometry}
\usepackage[numbers,sort]{natbib}
\usepackage{wrapfig}
\usepackage{booktabs}
\usepackage{tabularx}
\usepackage{enumitem}
\usepackage{floatrow}
\usepackage{xcolor}
\usepackage{soul}
\usepackage{caption}
\usepackage{subcaption}
\usepackage{caption}
\usepackage{lipsum}
\usepackage{fixltx2e}
\usepackage{graphicx}
\usepackage{hyperref}
\usepackage{amsmath}
\usepackage{authblk}
\usepackage{bbm}
\usepackage{amssymb}
\colorlet{lred}{red!40}
\colorlet{lgreen}{green!40}
\colorlet{lblue}{blue!40}
\definecolor{bananamania}{rgb}{0.98, 0.91, 0.71}

\usepackage[utf8]{inputenc}
\usepackage[T1]{fontenc}
\usepackage{fontenc}

\usepackage{siunitx}
\usepackage{float}
\restylefloat{figure}

\DeclareMathOperator*{\argmin}{arg\,min}
\DeclareSIUnit[number-unit-product = ]\percent{\char`\%}

\numberwithin{equation}{section}
\numberwithin{table}{section}
\numberwithin{figure}{section}

\usepackage[switch,columnwise]{lineno}
\newfloatcommand{capbtabbox}{table}[][\FBwidth]

\begin{document}
	\let\WriteBookmarks\relax
	\def\floatpagepagefraction{1}
	\def\textpagefraction{.001}
	

	\newcommand{\com}{\color{red}}
	\newcommand{\scheiss}{\color{green}}
	\newcommand{\riskpar}{\mathcal R^\text{par}}
	\newcommand{\riskblc}{\mathcal R^\text{blc}}
	\newcommand{\blc}{B}
	\newcommand{\dist}{P}
	\newcommand{\ali}{A}
	\newcommand{\rgb}{I}
	\newcommand{\net}{\mathcal N}
	\newcommand{\wass}{\mathcal W}
	\newcommand{\loss}{\ell}
	\newcommand{\R}{\mathbb{R}}
	\newcommand{\cspace}{[0,255]^{m_1\times m_2\times 3}}

	\newcommand{\abs}[1]{\left\lvert#1\right\rvert}
	\newcommand{\norm}[1]{\left\lVert#1\right\rVert}
	\newcommand{\hlf}[2]{\sethlcolor{#1}\hl{#2}}
	\newcommand{\alle}[2]{{\hlf{yellow}{\textbf{[#1@all: #2]}}}}
	\newcommand{\lec}[2]{{\hlf{lred}{\textbf{[#1@LEC: #2]}}}}
	\newcommand{\uibk}[2]{{\hlf{lgreen}{\textbf{[#1@UIBK: #2]}}}}
	\newcommand{\adela}[2]{{\hlf{lgreen}{\textbf{[#1@ADELA: #2]}}}}
	\newcommand{\christoph}[2]{{\hlf{lgreen}{\textbf{[#1@CHRISTOPH: #2]}}}}
	\newcommand{\markus}[2]{{\hlf{bananamania}{\textbf{[#1@M: #2]}}}}
	\newcommand{\steinbjoern}[2]{{\hlf{lblue}{\textbf{[#1@STEINBJOERN: #2]}}}}
	\newcommand{\innio}[2]{{\hlf{lblue}{\textbf{[#1@INNIO: #2]}}}}
	\newcommand{\mc}[1]{\mathcal{#1}}
	\newcommand{\jenbacher}{INNIO Jenbacher GmbH \& Co OG }
	\newcommand{\nparam}{\mc N_\theta^\text{par}}
	\newcommand{\nblc}{\mc N_\rho^\text{blc}}
	\newcommand{\lparam}{\ell^\text{par}}
	\newcommand{\lblc}{\ell^\text{blc}}
	\newcommand{\sk}[1]{\text{Sk}\left(#1\right)}
	\newcommand{\vvv}[1]{\text{Vvv}\left(#1\right)}
	\newcommand{\vmp}[1]{\text{Vmp}\left(#1\right)}

	\newcommand{\redhl}[1]{\textbf{\textcolor{red}{#1}}}
	
	\setlength{\emergencystretch}{2em}
	
	\newcommand{\abbreviations}[1]{%
		\nonumnote{\textit{Abbreviations:\enspace}#1}}
	
	\providecommand{\keywords}[1]{\textbf{\textit{Keywords---}} #1}
	
	\title {Surface Topography Characterization Using a Simple Optical
		Device and Artificial Neural Networks}

	\author[1]{Christoph Angermann}
	\author[1]{Markus Haltmeier}
	\author[2]{Christian Laubichler}
	\author[3]{Steinbj\"orn J\'{o}nsson}
	\author[1]{Matthias Schwab}
	\author[1]{Ad\'{e}la Moravov\'{a}}
	\author[2]{Constantin Kiesling}
	\author[2]{Martin Kober}
	\author[3]{Wolfgang Fimml}
	
	\affil[1]{Department of Mathematics, University of Innsbruck, Technikerstraße 13, \hspace{10cm} 6020 Innsbruck, Austria
		\hspace{10cm}
		\url{applied-math.uibk.ac.at}}
	
	\affil[3]{INNIO Jenbacher GmbH \& Co OG, Achenseestrasse 1-3, 6200 Jenbach, Austria\hspace{10cm}\url{www.innio.com/en}}

	\affil[2]{LEC GmbH, Inffeldgasse 19, 8010 Graz, Austria
		\hspace{10cm} \url{www.lec.at}}

	\maketitle

\begin{abstract}
	State-of-the-art methods for quantifying wear in cylinder liners of large internal combustion engines 
	require disassembly and cutting of the 
	liner. This is followed by laboratory-based high-resolution microscopic surface depth measurement that quantitatively evaluates wear based on bearing load curves 
	(Abbott-Firestone curves). Such methods are destructive, time-consuming and costly. The goal of the research presented is to develop nondestructive yet reliable methods for quantifying the surface topography. A novel machine learning framework is proposed that allows prediction of the bearing load curves from RGB images of the liner surface that can be collected with a handheld microscope. A joint deep learning approach involving two neural network modules optimizes the prediction quality of surface roughness parameters as well and is trained using a custom-built database containing 422 aligned depth profile and reflection image pairs of liner surfaces. The observed success suggests its great potential for on-site wear assessment of engines during service.
\end{abstract}


	\maketitle






\section{Introduction}

Digitalization offers a wide range of tools for large internal combustion engines, which are used in a variety of applications such as power generation or transportation. While digitalization in this field is not new, data-driven methods based on artificial intelligence (AI) and in particular machine learning (ML) have opened up promising new avenues. Although the origins of these digital technologies date back to the 1950s and even earlier \citep{carbonell1983,goodfellow2016}, their recent rise is due to the easy availability of data and the technical capabilities to process it efficiently \citep{mittal2019}.

In the field of large engines, data-driven methods are highly valuable in applications like condition monitoring (CM) and predictive maintenance (PdM) \citep{carvalho2019,zhang2019,lei2020}. With the help of data-driven and hybrid models (the latter combines first principles knowledge with knowledge derived from the data itself), CM and PdM concepts have the potential to avoid out-of-spec function, unforeseen downtime and premature component replacement
\citep{cartalemi2019,teichmann2019,jimenez2020,tiddens2020}, thereby increasing key component lifetime and reducing the engine’s carbon footprint \citep{basurko2015}. Yet the generation of sufficient high-quality data as a reliable basis for such data-driven models can be challenging.
This is especially true for wear-related data generation, the focus of this paper. Up-to-date CM and PdM concepts for cylinder liners frequently build upon indirect methods such as lubricating oil or vibration condition monitoring \citep{rao2022}. However, continuous direct and position-based measurement of wear of cylinder liners is currently not possible during engine operation. Even at engine standstill, measurement is elaborate or even unfeasible. For large internal combustion engines for stationary power generation, the liner must be disassembled and cut before surface depth profiles for wear assessment can be measured with a sophisticated optical measurement device, making component reuse and further wear assessments at later stages impossible.

\subsection{Cylinder Liner Wear}

The inside of the cylinder liner serves as both a tribological contact surface and a sealing surface for the piston and the piston rings. Its desired surface properties are achieved by a manufacturing process known as honing. In this process, the surface is formed into a fine plateau structure that simultaneously minimizes friction and facilitates oil retention. 
The cylinder surface is honed by rotating a device with abrasive stone inserts while moving the device vertically. 
Such plateau honing is a conventional honing method used for cylinders in the gas engine industry \citep{hrdina2020}. The surface finish is an important factor for oil consumption, piston-ring friction, wear propensity and blow-by of cylinders \citep{hill1995,sato2004,schmid2005}. In addition, the cylinder liner surface topography also affects the engine emissions \citep{schmid2013,tomanik1992}.


Due to the movement of the piston during the combustion cycle, the inner surface of the cylinder liner is subject to constant wear, which alters important surface properties. Besides mechanical friction, the honing structure can also wear out due to intrusion of foreign particles from either oil coking or fuel contamination \citep{wakuri1988,ye2004,greuter2012}. As a consequence, surface wear decreases the volumetric efficiency of the engine and increases blow-by; oil consumption; power loss and HC, CO, CO\textsubscript{2} and NO\textsubscript{x}-emissions \citep{korcek1999,patel2015,rahnejat2010}. In addition, wear impairs the hydrodynamic support of the piston rings and thus increases the risk of a fatal engine failure \citep{obara2016}.

Comprehensive simulation work and component inspections at \jenbacher on gas engines have established that the greatest wear occurs near top dead center (TDC). In this region, the cylinder liner wall temperature is comparatively high due to the thermal impact from combustion. This leads to low oil viscosity, a thin oil film and thus asperity contacts between the compressed piston ring and the cylinder wall \citep{henein1997}. The continuous movement of the piston causes a change in the thrust force in the cylinder and also a continuous change in both the circumferential and the vertical directions, namely from TDC to bottom dead center (BDC). As a result, greater wear occurs in the liner surface area parallel to the piston pin axis on the thrust/anti-thrust side of the cylinder, while less wear is observed in the area perpendicular to the pin axis (see Figure \ref{fig:wear}).
Due to the peak firing pressure in the cylinder, an additional contact pressure arises between the piston ring and the liner, which further increases wear at the liner areas near TDC.
In general, the greatest wear occurs in the area parallel to the piston pin axis near TDC, while areas perpendicular to the piston pin axis near BDC are considered unworn.

\subsection{Surface Roughness Characterization}

\begin{figure*}[htb!]
	\centering
	\begin{floatrow}
	\ffigbox[.38\columnwidth]{
		\includegraphics[width=0.5\columnwidth]{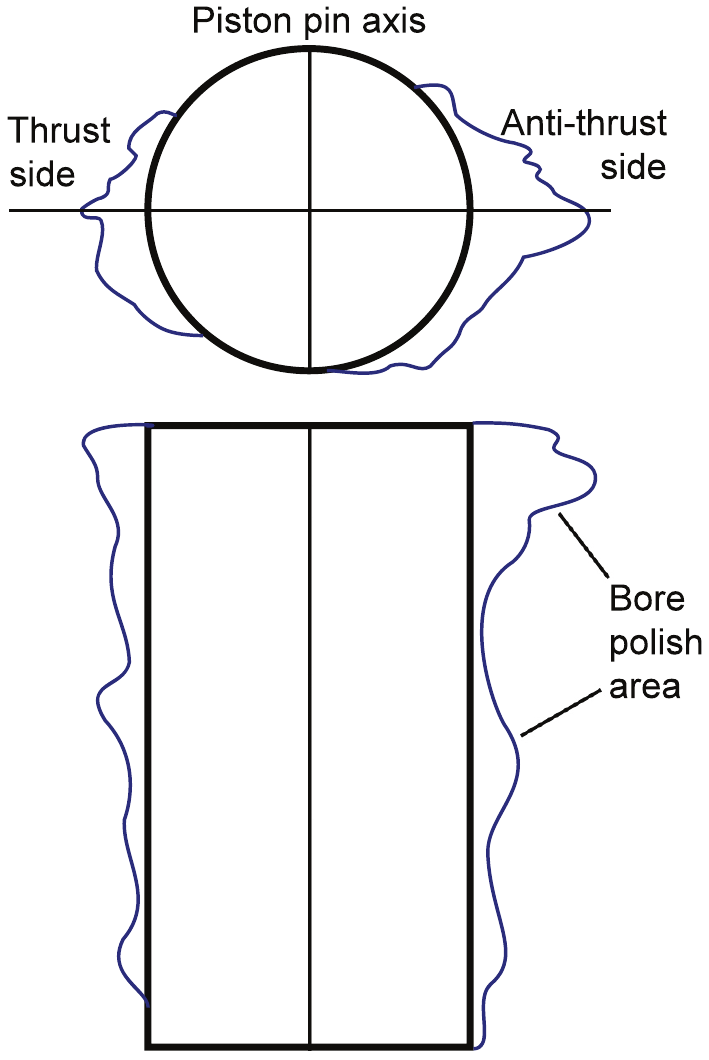}}{
		\caption{Example of the {observed} circular and vertical wear distribution of a cylinder liner in operation. The density function describes the amount of wear accumulated at the cylinder wall.} \label{fig:wear}}\hspace{1em}
	\ffigbox[.57\columnwidth]{
		\includegraphics[width=1.01\columnwidth]{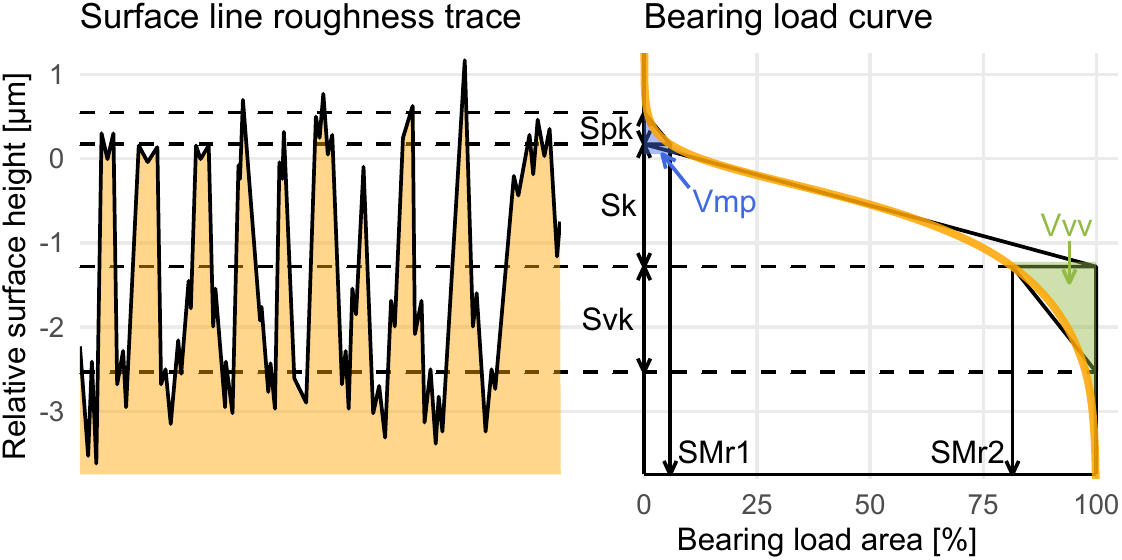}}{
	\caption{\label{fig_params} The bearing load curve (also known as the Abbott-Firestone curve) of a surface depth measurement. The corresponding functional roughness indicators Sk, Spk, Svk, SMr1, SMr2 and volume parameters Vvv, Vmp are derived directly from the BLC depth representation \citep{salcedo2018,bigerelle2007}.}}
\end{floatrow}
\end{figure*}

Evaluation of the surface roughness and the lubrication characteristics of cylinder liners can help to determine signs of failure or premature wear. A common approach in the large engine industry is to evaluate wear by taking spatial depth measurements of an unworn and a worn area of the cylinder liner and then comparing the corresponding bearing load curves (BLC) after height compensation \citep{obara2016,kumar2000,mezghani2013,cabanettes2014,laubichler2021}. The BLC is also known as the bearing surface curve or Abbott-Firestone curve. Mathematically speaking, the BLC is defined as the reversed empirical quantile function of the observed depth profile values. As illustrated in Figure \ref{fig_params}, the BLC plots the reduced height value against the percentage of measurement points above that height. In terms of cylinder liner functionality, the BLC describes two important parts of the honing structure: the valley part, which specifies the oil retention capacity, and the peak part, which specifies debris or asperities of the honing structure \citep{dong1995}. Depending on the measurement method, either profile or surface parameters are visualized by the BLC.

Profile values (denoted with an R prefix) are determined using conventional stylus profilometers \citep{mathia2011}. These devices measure the surfaces with a needle that traverses a defined distance and outputs the corresponding BLC. Surface values (denoted with an S prefix) are determined by using optical methods. Unlike stylus instruments, optical methods are able to record the entire surface and are less sensitive to measurement errors. Depending on whether profile or surface measurements are used, commonly extracted functional parameters from the BLC are:
\begin{itemize}
	\itemsep0em 
	\item heights of core roughness areas Rk and Sk
	\item reduced peak heights Rpk and Spk
	\item reduced valley heights Rvk and Svk 
	\item material ratios of peaks  RMr1 and SMr1
	\item material ratios of valleys RMr2 and SMr2.
\end{itemize}
A definition of and detailed information on parameter calculation out of the BLC are given in ISO 13565-2 (profiles) and ISO 25178 (surfaces) and further discussed in \citep{salcedo2018,bigerelle2007}.
The set of  volume parameters also introduced in ISO 25178 standard is  of high relevance for curve characterization but less frequently used:
\begin{itemize}
	\itemsep0em 
	\item peak material volume Vmp
	\item valley void volume Vvv
	\item core material volume Vmc
	\item core void volume Vvc.
\end{itemize}

The spatial measurements that allow for calculation of a BLC are usually performed with an advanced optical device that generates a high-resolution microscopic surface depth profile. Since the surfaces of interest to the large engine industry are located on the inside of the cylinder liner and the optical device is attached to a fixed measurement station due to its cumbersome size, it is necessary to disassemble the cylinder liner and cut it into segments containing the respective surface areas.
 A specific issue with this method is that the components are destroyed and cannot be used again after measurement, i.e., a single component cannot be measured at a later stage when it has accumulated more wear. In addition, handling the sophisticated measurement equipment requires a high level of technical understanding of roughness assessment and software-specific training on the measurement device. In summary, the current measurement pipeline is a time and resource-intensive process chain, which makes it unsuitable for constant evaluation of production quality and long-term studies of the wear of cylinder liner components.

\subsection{Machine Learning for Surface Characterization}
The goal of the research presented here is to use fast yet reliable and informative methods in combination with advanced machine learning algorithms to evaluate the wear condition of engine components without their removal and destruction. Current state-of-the-art technology integrates optical sensors and especially digital image sensors into comparatively small and inexpensive devices such as cell phones, compact cameras and handheld microscopes. These devices are capable of producing a large amount of image data in a time-saving, simple and non-invasive manner \citep{burggraaff2019,kanchi2018}. The challenge is to make a reliable prediction based on this data, since an RGB modality does not directly describe actual surface depth, it offers contextual information about the surface.

In general, machine learning-based systems acquire their knowledge from high-dimensional raw data \citep{goodfellow2016}. The use of neural networks with a high amount of network layers is referred to as deep learning and enables the learning of complex data structures by using multiple levels of simpler, more abstract representations \citep{goodfellow2016,lecun2015}.
  In modality transfer learning, observed data of a simpler modality in terms of acquisition effort is mapped to desired properties of a more complex modality. In the context of cylinder liner wear assessment, this means that the BLC is not derived from a sophisticated depth measurement but predicted by a simpler RGB modality. A variety of existing methods predict individual roughness parameters with machine learning-based or even deep learning-based approaches \citep{rifai2020,lawrence2014}. Rifai et al. \citep{rifai2020} modeled the average roughness without predicting any information about the condition of the valleys and the peaks of the surface. Lawrence et al. \citep{lawrence2014} were able to give quite reliable predictions on the Sk, Svk, Spk, SMr1, SMr2 parameters extracted from the BLCs, using a stationary vision system with a sophisticated CCD sensor.
However, there is no approach that focuses on the prediction of the entire BLC curve out of simple RGB data obtained by a handheld device.

In this research, a novel joint deep learning framework is trained from easy-to-obtain and low-resolution RGB reflection images of the inner surfaces of gas engine cylinder liners to finally predict the BLC of the depth profile. First, the input image is propagated through a preprocessing pipeline using multiple high-pass filters to extract one-dimensional representations of the input.
In the parameter prediction module, three surface parameters that describe the BLC shape well are predicted from the processed signals using a deep fully-connected neural network. In the BLC prediction module, these predicted parameters are combined with the processed signals and propagated through a convolutional neural network (CNN) to predict the overall BLC. CNNs represent the state of the art for several image processing tasks such as segmentation \citep{long2015,angermann2021}, image reconstruction \citep{gupta2018,yao2019} or modality transfer  \citep{han2017, eigen2014, lei2019,godard2017}. CNNs are also gaining popularity in the field of 1D signal processing, especially in combination with recurrent networks and time series data \citep{malek2018,karim2018,kiranyaz2021}.

For this data-driven prediction methodology, a database is built from scratch. 
For ground truth data, engine liners are removed and 3D depth profiles are generated using a confocal microscope. These profiles of worn and unworn areas are used to calculate the corresponding BLCs.
Reflectance RGB images of the same areas are acquired via a handheld digital microscope, which exemplifies the wide range of simple handheld optical methods that are available. Once the network is trained, the BLC of a new liner can be predicted from a simple RGB image of the surface. This methodology for targeted modality transfer promises tremendous improvement in terms of time resources, replacement liner costs and measurement costs since it eliminates the need to remove and destroy cylinder liners for measurement with expensive and sophisticated microscopy equipment. The main focus of this study is inner surface inspection of cylinder liners in large gas engines. Nevertheless, it should be emphasized that the proposed machine learning framework can be similarly applied to other image-based inspection tasks.

\subsection{Outline}
The present paper is organized as follows. Section \ref{sec:data} describes the considered optical modalities and introduces a self-built database used to train and evaluate the proposed modality transfer learning framework for nondestructive and accelerated wear assessment. Section \ref{sec:methodology} presents the framework in detail. Section \ref{sec:results} contains results and discussions of the proposed BLC prediction procedure. Finally, Section \ref{sec:conclusion} summarizes the main findings of this study and discusses important future research directions.


\section{Data}
\label{sec:data}

\subsection{Optical Modalities}
\label{sec:sub-mod}

Existing wear analysis techniques require a complex workflow to generate high-resolution depth measurements. The depth modality is replaced by optical reflection images combined with a modality transfer approach. The specific image modalities are described below.

\paragraph{Fixed confocal device outputs:}
The Alicona InfiniteFocus confocal microscope \citep{alicona} used in this work provides a depth image of a given measurement area. These cylinder liner surface depth images on the nanoscale are measured in a spatial domain of approximately \SI[]{1.9x1.9}{\mm} with a resolution of \num{3167x3158} pixel. Each pixel describes the distance of the surface point from the normalized surface height.

\paragraph{Handheld device outputs:}

\begin{figure}[htb!]
	\begin{floatrow}
		\ffigbox[.99\columnwidth]{
			\includegraphics[width=0.18\columnwidth]{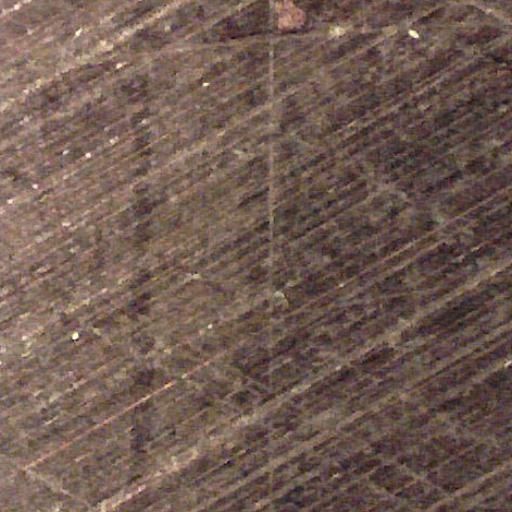}\hfill
			\includegraphics[width=0.18\columnwidth]{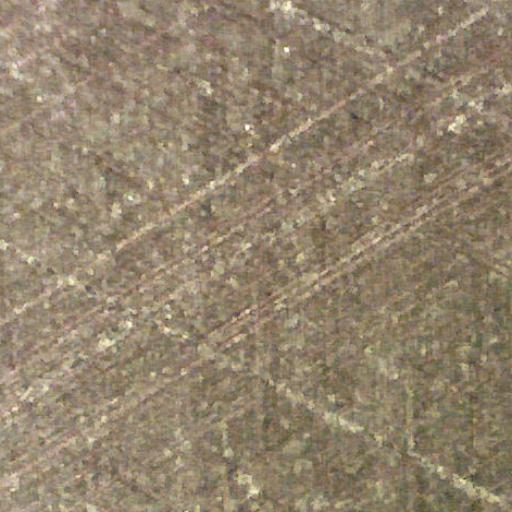}\hfill
			\includegraphics[width=0.18\columnwidth]{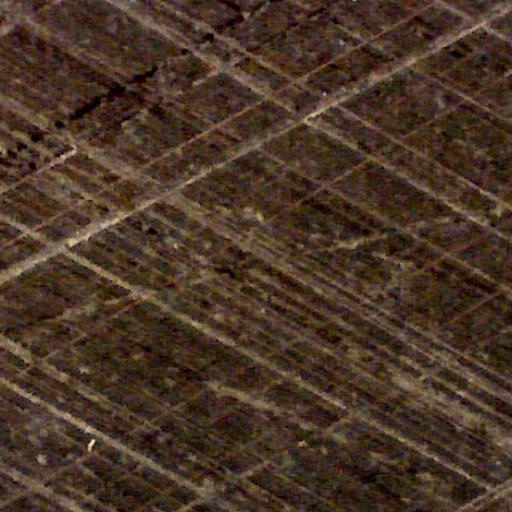}\hfill
			\includegraphics[width=0.18\columnwidth]{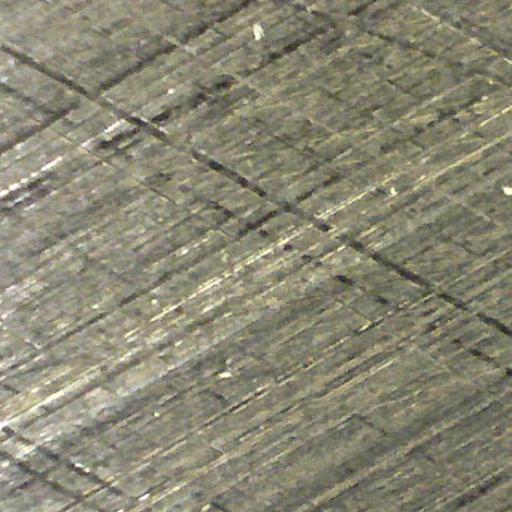}\hfill
			\includegraphics[width=0.18\columnwidth]{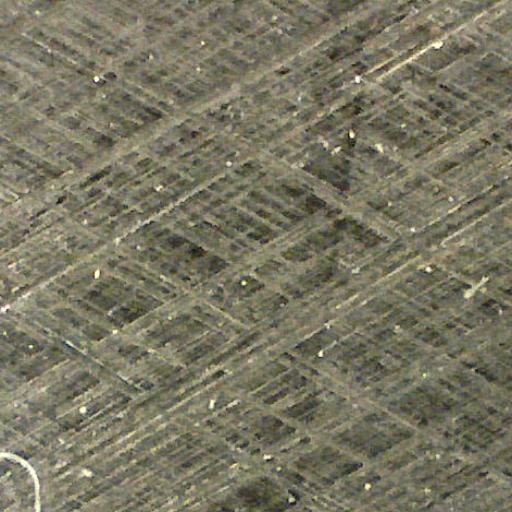}}{
			\captionof{figure}{Examples of RGB measurements of distinct cylinder liners, recorded with $80 \times$ magnification using a handheld Mic-Fi digital Wi-Fi microscope \citep{micfi}.}
			\label{fig:micfi}}
	\end{floatrow}
\end{figure}

Of great practical relevance is the optical surrogate modality. A Mic-Fi digital Wi-Fi microscope  \citep{micfi} with a length of \SI{135}{\mm} captures RGB reflection images in SXGA format (\num{1280 x 1024} pixels) at up to 220 times the magnification. Measurements cover a region of approximately \SI[]{4.2 x 4.2}{\mm}.
Such a handheld device can potentially inspect cylinder liners without full disassembly of the component, which facilitates inner surface imaging of a permanently mounted liner.
Figure \ref{fig:micfi} visualizes images taken with the handheld device. It highlights additional practical challenges of using this handheld device such as low resolution, non-uniform light intensities, artifacts, noise, blurred areas and different image quality. \\

The transfer learning model presented below is not limited to this specific handheld device, which is used here mainly because this microscopic camera fits well with the cylinder dimensions. Any other handheld device for RGB image acquisition such as a compact camera, smartphone or endoscope could be used instead.

\subsection{Database Generation}
\label{ssec:database}

\begin{figure*}[htb!]
	\centering
	\includegraphics[width=0.9\textwidth]{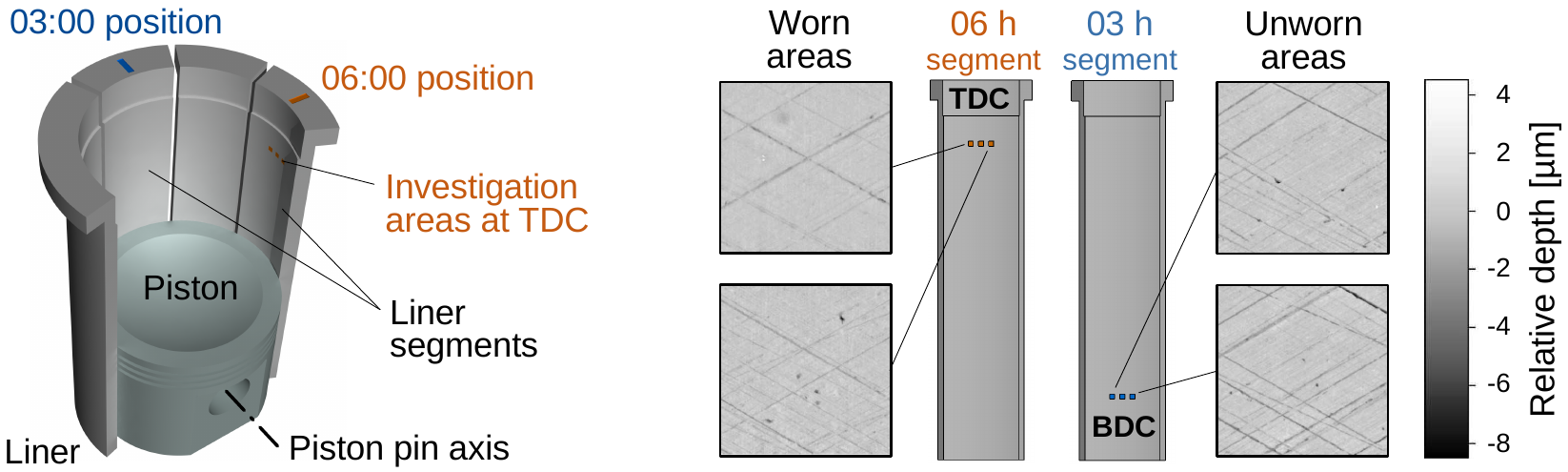}
	\caption{Left: A segment of \SI{45}{\degree} is cut out of each liner which contains the area where the greatest wear is expected (in this case the \SI{6}{\hour} position). Another segment is cut out at the \SI{3}{\hour} position where no wear is expected in the BDC. Right: Comparison of worn and unworn areas. The grayscale on the topographic scans represents the relative depth, i.e. the distance of the surface from the normalized surface height. The scans of the worn areas show less contrast due to the abrasions of the peaks.} \label{fig:martin}
\end{figure*}

In this study, a database was created from scratch. Ten type 6 gas engines from \jenbacher were examined. Each engine consists of 12, 16, 20 or 24 cylinders with a displacement of approximately \SI{6}{\cubic\deci\metre}. The number of operating hours of the corresponding cylinder liners varies between  $\SI{2550}{\hour}$  and $\SI{60000}{\hour}$; this number may differ within an engine due to past replacements of individual liners.  Each liner is inspected following a time-consuming and resource-intensive logistics chain, which initially consists of disassembling the liner, removing it from the engine and marking it. From each liner, two segments of $\SI{45}{\degree}$ are cut that contain positions parallel (\SI{6}{\hour}-segment) and perpendicular (\SI{3}{\hour}-segment) to the piston pin axis. The segments and measurement areas are illustrated in Figure \ref{fig:martin}. 
These segments are superficially cleaned and the confocal microscope measures three to five different areas within TDC of the \SI{6}{\hour}-segments and BDC of the \SI{3}{\hour}-segments.

\begin{figure}[htb!]
	\centering
	\begin{subfigure}{.365\columnwidth}
		\includegraphics[width=0.45\columnwidth]{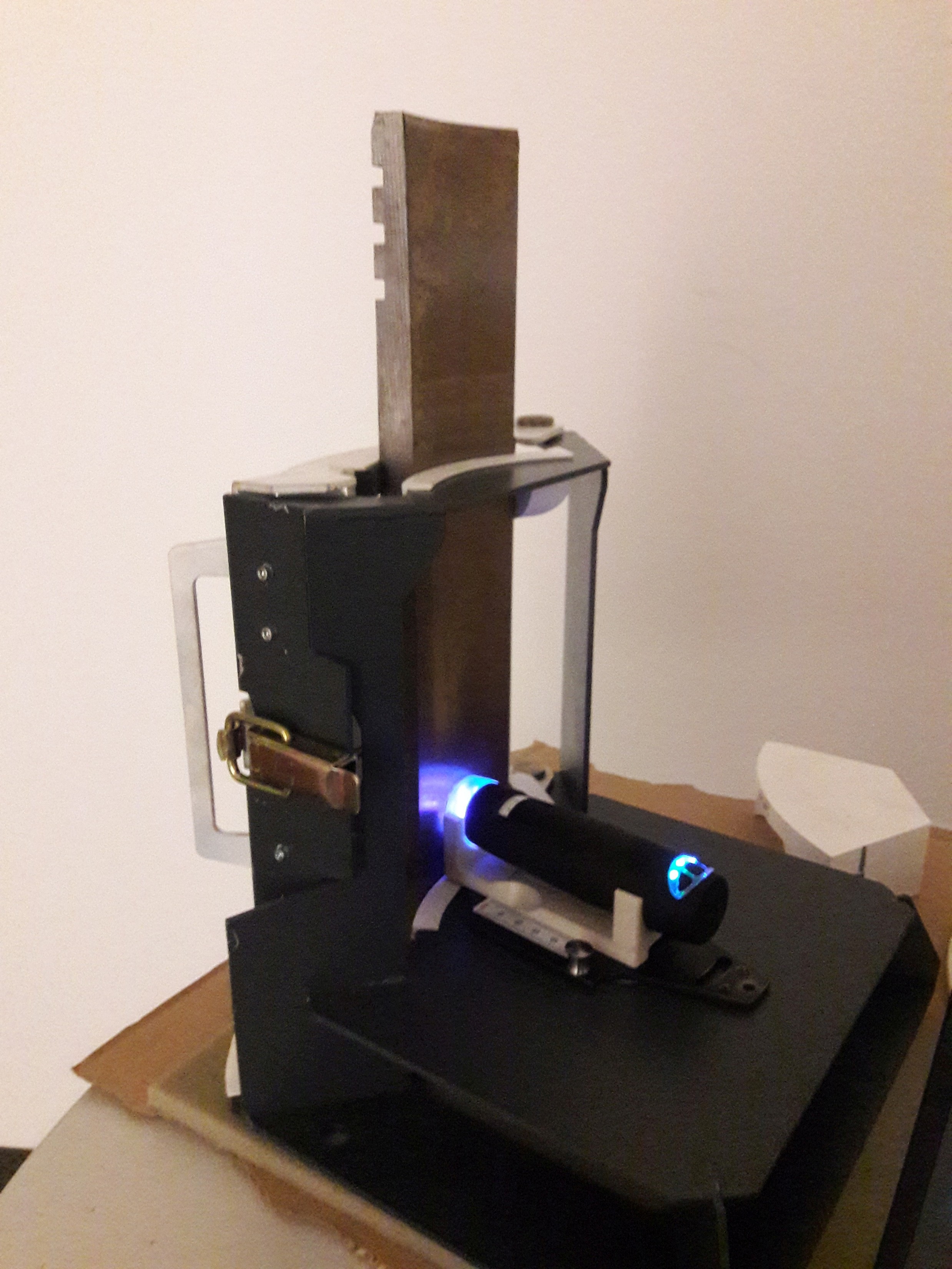}\hfil
		\includegraphics[width=0.45\columnwidth]{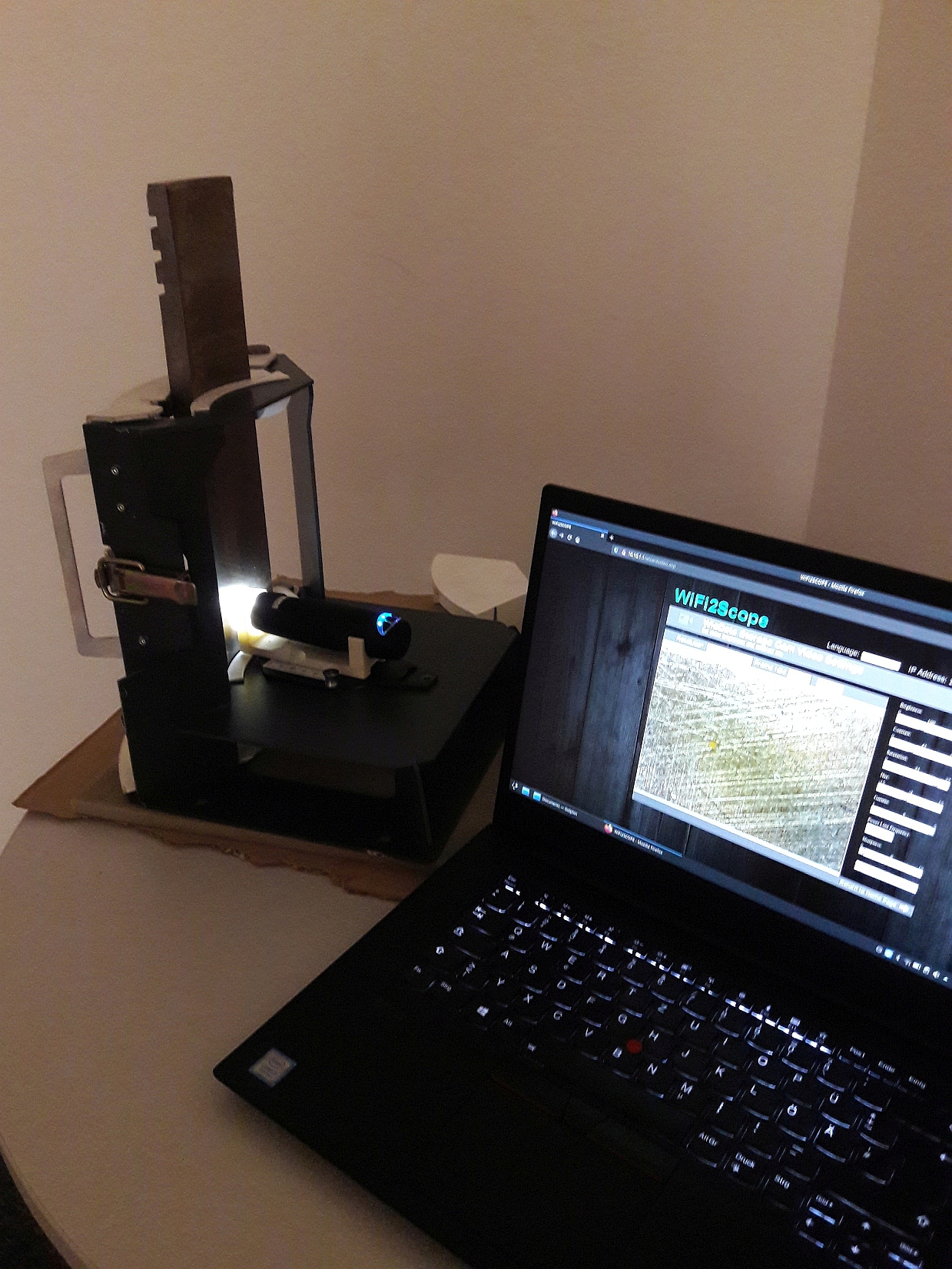}
	\end{subfigure}\hfill
	\begin{subfigure}{.585\columnwidth}
		\includegraphics[width=1.0\columnwidth]{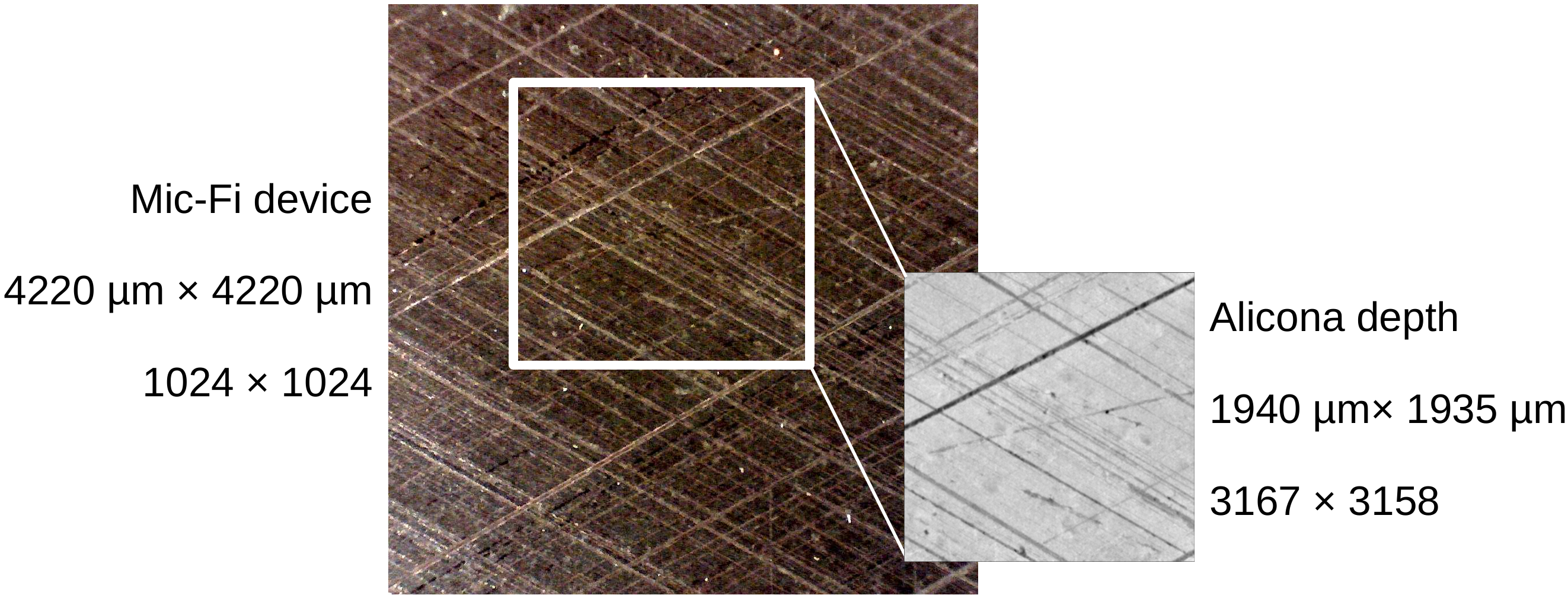}
	\end{subfigure}
	\caption{Left: Measurement station with a self-constructed segment holder. The holding device enables measurement of comparable positions relative to the confocal microscope. The camera transfers the recorded measurement via a Wi-Fi connection to mobile work stations. Right: Depth profiles from the fixed confocal device for BLC calculation are fully contained in the corresponding handheld microscope measurements. A subsequent registration step yields perfectly aligned data pairs.}
	\label{fig:holder}
\end{figure}

After the depth images are generated, the segments are forwarded to a second distinct measurement station, where approximately the same areas  are recorded with the handheld device using a self-constructed segment holder (Figure \ref{fig:holder}, left) placed in a darkened room. Due to very fine scaling on micrometer range, it is not possible to measure the exact same spatial range of the segment surface as the depth profile with the handheld device. The segment holding device ensures that the areas of the confocal microscope measurements are embedded in the corresponding handheld device records (Figure \ref{fig:holder}, right).  
After measurement, a postprocessing step is added in which the depth image sections are registered within the spatially larger handheld images. To be more precise, a translation of each measurement area is found that maximizes the mutual information between the depth profile and the RGB section. In addition, the RGB images are cropped to a size of \num{470 x 470} pixels to cover the same spatial area of \SI[]{1.9x1.9}{\mm} as the registered depth sections. The result is that perfectly aligned image sections of the handheld and the confocal microscope measurements are achieved, which subsequently allows the development of models in a supervised setting.
A total of 174 liner segments with 3 to 5 measurement areas per segment is given. Each registered modality pair is checked visually and removed if the registration fails, which finally results in a set of 422 modality pairs.

\section{Methodology}
\label{sec:methodology}

This section presents the proposed framework for nondestructive and fast surface topography characterization of cylinder liners based on surrogate image acquisition methods and machine learning. 

\subsection{BLC Prediction and Modality Transfer}

The goal is to replace the depth profile measurements with the simpler modality described above, which does not contain any information about pixel-wise depth, and to predict the BLC of the depth profile measurement from this indirect contextual RGB information.
Let $\ali  \in\R^{d_1\times  d_2}$ denote a recorded depth profile where $d_1\times  d_2$ denotes  the number of sampling points (pixels). In this paper, the BLC (the reversed empirical quantile function of the depth values) of a profile $\ali$ at position $x\in(0,1)$ is modeled as follows: 
\begin{equation}
\label{def:blc}
\mc B\colon\R^{d_1\times  d_2}\times (0,1)\to \R,\quad (\ali,x ) \mapsto  \inf\bigg\{ y \in\R\ \big|\  1 - x \leq \frac{1}{d_1 \cdot d_2}\sum_{A} \mathbb{1} \{a_i \leq y\}\bigg\} \ .
\end{equation}
Here the sum runs over all pixels $a_i$ in  profile $A$ and $ \mathbb{1} \{ a_i  \leq y \}$ denotes the indicator function, returning 1 if $a_i \leq y$ and 0 otherwise.
In fact, the discretized BLC $\blc(A) \in\R^K$ is considered, obtained after sampling the BLC $ \mc B$ in (\ref{def:blc})  at  $K$ equidistant sampling positions in the second argument, i.e., 
\begin{linenomath*}
	\begin{equation}
		\blc(A)\triangleq\left\{\mc B\left(A,\frac{k}{K+1}\right)\right\}_{k=1}^K.
	\end{equation}
\end{linenomath*}
 This allows the entire curve to be represented in a one-dimensional vector of size $K$ in subsequent implementations.  \\

Given an RGB reflection image $\rgb \in [0,255]^{m_1\times m_2 \times 3}$, the task consists of predicting BLC $\blc(A_I)$ derived from the corresponding depth image $\ali_I$ of the same area. Here $m_1\times  m_2$ is the number of image pixels and $3$ the number of color channels.  In mathematical terms, the task is to find a transfer function 
\begin{linenomath*} \begin{equation} \label{eq:transfer}
\mc T  \colon [0,255]^{m_1 \times m_2 \times 3} \to \R^K,\ \rgb \mapsto \blc(A_I)  \,,
\end{equation} \end{linenomath*}
 that maps the RGB reflection image onto the corresponding  BLC.  
Preliminary data analyses have suggested that RGB reflection images of a cylinder liner surface do indeed provide sufficient information to distinguish between light and severe wear. This supports the assumption that RGB reflection images permit the determination of the corresponding BLC of the depth image. It is worth noting that the prediction of the empirical quantile function of a different modality can be seen as a variant of modality transfer, in which low-dimensional information of a desired measurement modality is synthesized from a simpler data acquisition technique \citep{eigen2014,godard2017,han2017,lei2019}.


\subsection{Data Preparation}
A closer look at Figure \ref{fig:micfi} indicates, that RGB data recorded with a handheld device exhibits disadvantages such as weak resolution, inconsistent illumination, artifacts, noise, blurred areas and different image quality. Therefore, instead of using the raw images, the image data is propagated through a preprocessing pipeline before it is processed by the prediction model. The following process is inspired by the idea of Lawrence et al. \citep{lawrence2014}, which enables the prediction of the Sk, Spk, Svk, SMr1 and SMr2 parameters out of high quality images recorded by a stationary vision system.

 In the preprocessing step, an input RGB image $\rgb \in \cspace$ is converted to grayscale using the standard luminosity method \cite{kanan2012}, i.e., $\rgb_\text{gray}\triangleq \frac{0.299}{255}\rgb_{(\cdot,\cdot,1)}+ \frac{0.587}{255}\rgb_{(\cdot,\cdot,2)}+ \frac{0.114}{255}\rgb_{(\cdot,\cdot,3)}\in \R^{m_1\times m_2}$. As described in \citep{lawrence2014}, both the illumination source and reflectance properties of the object contribute to the image information, where illumination is usually characterized by low frequencies and the reflectance component by higher frequencies. It is desirable to eliminate the illumination source in the handheld RGB data for the subsequent depth prediction. Therefore, $\rgb_\text{gray}$ is convolved with high-pass filters using different cut-off frequencies. Varying cut-off frequencies are used due to the varying illumination and quality in the recorded RGB data. A high-pass filter with cut-off frequency distance $\sigma$ from the origin is applied by multiplication of the 2D Fourier transform of $\rgb_\text{gray}$ with Gaussian filter matrices $H^\sigma\in[0,1]^{d_1\times d_2}$ defined by 
 \begin{linenomath*}
 	\begin{equation}
		 H^\sigma_{i,j}\triangleq 1-\exp\left(-\frac{\left(i-d_1/2\right)^2+\left(j-d_2/2\right)^2}{2 \sigma^2}\right) \quad   \text{for}\  i=1,\ldots,d_1\  \text{and}\  j=1,\ldots,d_2. 
 	\end{equation}
 \end{linenomath*}
 This results in the filtered outputs $\rgb_\text{gray}^\sigma$, where four different values for $\sigma$ are considered: $8,16,32,64$. In fact, the BLCs of the filtered profiles are considered as input for the subsequent data-driven approach. Taken all together, the transformation module is described by
 \begin{linenomath*}
\begin{equation}
\label{eq:prepro}
\Psi\colon\cspace\to \R^{K\times 4},\quad \rgb \mapsto \left[\blc\left(\rgb_\text{gray}^8\right),\blc\left(\rgb_\text{gray}^{16}\right),\blc\left(\rgb_\text{gray}^{32}\right),\blc\left(\rgb_\text{gray}^{64}\right)\right].
\end{equation}
\end{linenomath*}

\subsection{Proposed Machine Learning Approach}
\label{ssec:approach}
A database is created that contains $N$ training data pairs $ (\rgb_n,\blc_n)$ of RGB input images $\rgb_n \in \cspace$ and corresponding BLCs $\blc_n \in \R^K $ for $n =1,\ldots,N$. The transfer function $\mc T$ \eqref{eq:transfer} for prediciting BLCs out of RGB images is composed of the preprocessing module $\Psi$ \eqref{eq:prepro} and a subsequent joint machine learning approach including the following modules:
\subsubsection{Parameter Prediction Module}
\label{sec:paramestim}
In the first module, meaningful BLC parameters are predicted in order to ensure a reasonable shape of the overall BLC in the final transfer function.  
Earlier investigations by the authors have shown that the BLC prediction accuracy of the whole system improves if the following three parameters are given as additional input besides the processed RGB input: the core roughness Sk, which indicates the flattest \SI{40}{\percent} slope of the BLC in the core region (\SI{20}{\percent}-\SI{70}{\percent}), the valley void volume Vvv, which depends on the last \SI{20}{\percent} of the bearing load area, and the peak material volume Vmp, which describes the first \SI{10}{\percent} of the bearing load area. The three parameters are calculated independently of each other. Thus the output is modeled as a multivariate response variable.\\

First, image-based parameters are calculated based on the BLCs of the preprocessed profiles $\Psi(\rgb_n)\in\R^{K\times 4}$ \eqref{eq:prepro}, which yields the input vector 
\begin{linenomath*}
	\begin{equation}
			p_n^I\triangleq  \left[\sk{\Psi(\rgb_n)}^T,\vvv{\Psi(\rgb_n)}^T,\vmp{\Psi(\rgb_n)}^T\right]\in \R^{3\cdot4}.
	\end{equation}
\end{linenomath*}
 Second, a class of deep fully-connected neural networks $\nparam \colon\R^{12} \to \R^3$ is designed, where $\theta$ denotes the model parameters and $\theta \mapsto \nparam$ is referred to as the network architecture.
  Third, a loss function $\lparam \colon \R^3 \times \R^3 \to [0, \infty]$ is constructed that measures the deviations between the module output $\nparam(p_n^{\rgb})$ and the ground truth parameters $p_n^B\triangleq\left[\sk{\blc_n}^T,\vvv{\blc_n}^T,\vmp{\blc_n}^T\right]$. Finally, mathematical optimization algorithms are employed to minimize the empirical risk function 
\begin{linenomath*}
\begin{equation} \label{eq:parrisk}
	\riskpar(\theta) \triangleq  \argmin_{ \theta}
	\frac{1}{N}\sum_{n=1}^{N}\lparam\left(\nparam\left(p_n^I\right),p_n^B\right)\,.
\end{equation}
\end{linenomath*}
The final trained module is given by $ \net^\text{par} = \net_{\hat\theta}^\text{par}$, where $\hat\theta$ is determined to be close to a global minimizer of the empirical risk function \eqref{eq:parrisk}. \\

Although the preprocessing pipeline and the parameter prediction approach were initially inspired by Lawrence et al. \citep{lawrence2014}, it is important to note that their approach clearly differs from the proposed framework here. Lawrence et al. used a single data-optimized high-pass filter due to consistent and high image quality measured with a stationary vision system and a CCD sensor. Furthermore, the subsequent prediction of the Sk, Svk, Spk, SMr1 and SMr2 parameters were based on a fully connected network with only one hidden layer. In this work, it is necessary to compensate for the comparatively poor quality of handheld RGB data by introducing several high-pass filters and a class of deeper neural networks.

\subsubsection{BLC Prediction Module}
\label{sec:blcestim}

\begin{figure}[htb!]
	\centering
	\includegraphics[width=.99\columnwidth]{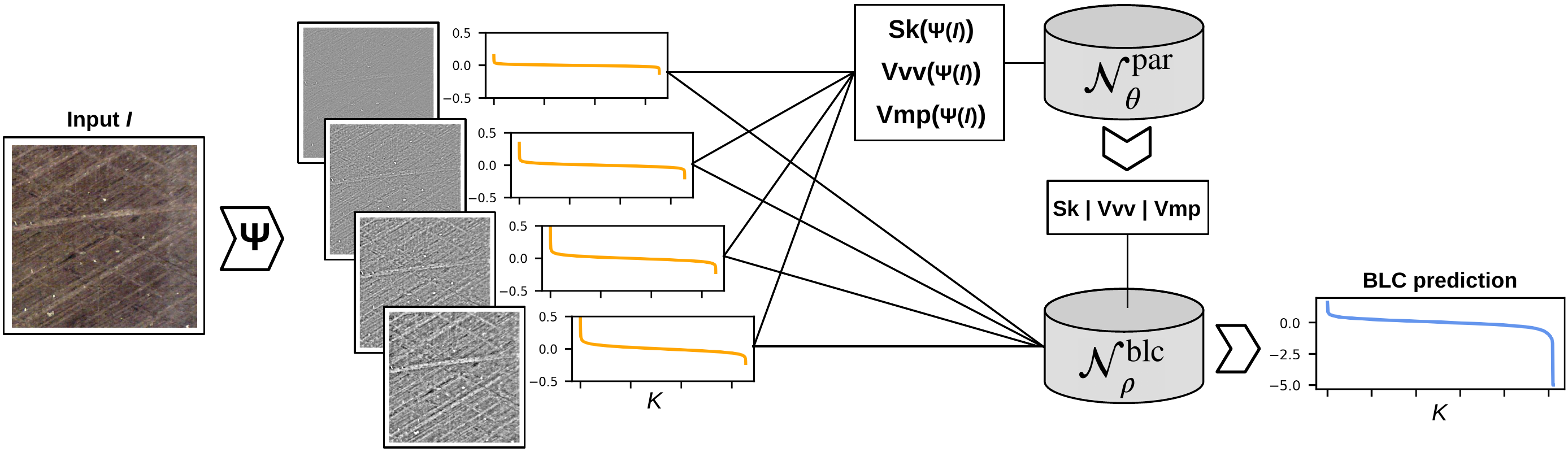}
	\caption{Transfer function $\mc T$ \eqref{eq:transfer} for BLC prediction. From left to right:  RGB sample $I$ recorded by hand is passed through the preprocessing module $\Psi$ \eqref{eq:prepro}, which converts the RGB to grayscale, applies four high-pass filters and calculates the reversed empirical quantile functions of these four filtered profiles. The Sk, Vvv and Vmp values, extracted from the processed curves, serve as an input for the parameter prediction module $\nparam$. The interim predicted Sk, Vvv and Vmp values are propagated together with the previous four input curves through the BLC prediction module $\nblc$, which yields the final prediction.}
	\label{fig:method}
\end{figure}

The input $\Psi(\rgb_n)\in \R^{K\times 4}$ can be seen as a one-dimensional signal of length $K$ with 4 different channels. The channel size is augmented to 7 by using the output from the  parameter prediction module, i.e., each entry of $\nparam(p_n^{\rgb})\in\R^3$ \eqref{eq:parrisk} is repeated $K$-times and concatenated with $\Psi(\rgb_n)$.
 Second, a class of CNNs $\nblc \colon\R^{K\times 7} \to \R^K$ is designed, where $\rho$ denotes the model parameters and $\rho \mapsto \nblc$ is referred to as the network architecture. 
The use of convolutional layers is motivated by the fact that input and output layers denote quite long signals and the units are not independent of adjacent units.
 Third, a loss function $\lblc \colon \R^K \times \R^K \to [0, \infty]$ is constructed that measures the deviations between the module output $\nblc\left(\mc C\left(\left[\Psi(\rgb_n),\nparam(p_n^{\rgb})\right]\right)\right)$ and the ground truth BLC $\blc_n$, where $\mc C$ denotes the concatenation via the channel dimension. Finally, mathematical optimization algorithms are used to minimize the empirical risk function: 
\begin{linenomath*}
	\begin{equation} \label{eq:blcrisk}
	\riskblc(\rho) \triangleq  \argmin_{ \rho}
	\frac{1}{N}\sum_{n=1}^{N}\lblc\left(\nblc\left(\mc C\left(\left[\Psi\left(\rgb_n\right),\nparam\left(p_n^{\rgb}\right)\right]\right)\right),\blc_n\right)\,.
	\end{equation}
\end{linenomath*}
The final trained module is given by $ \net^\text{blc} = \net_{\hat\rho}^\text{blc}$, where $\hat\rho$ is determined to be close to a global minimizer of the empirical risk function \eqref{eq:blcrisk}. The entire transfer function $\mc T$ \eqref{eq:transfer}, consisting of the preprocessing pipeline and the subsequent joint machine learning approach, is graphically summarized in Figure \ref{fig:method}. 

\subsection{Module Architectures}
\label{ssec:architecture}
For the first module, the parameter prediction part, a family of neural networks with five fully-connected (dense) layers is considered and described in detail in Table \ref{tab:nparams}. In the following tabular summaries, \textbf{input} corresponds to the input of each layer, \textbf{activ} the activation function followed after each layer and \textbf{\#weight} the number of free parameters of each layer. Network input and output are denoted by $\mc I$ and $\mc O$, respectively. 

In the second module, deployment of CNNs is proposed for BLC prediction. The key component of CNNs are convolutional layers, where each channel of the input is convolved with several data-adapted kernels. For each kernel, convolution with the layered input followed by application of a nonlinear activation function results in hidden feature maps representing properties of the input channels on different scales. To train a CNN, the most informative feature maps are found by fitting convolution kernels to the given data. In the proposed BLC prediction approach (cf. Table \ref{tab:nblc}), each convolutional layer (conv) is followed by an instance normalization \citep{ulyanov2018}.

\begin{table}[h!]
	\begin{floatrow}

	\ttabbox[.46\columnwidth]{\caption{Module architecture ${\nparam}$. \textbf{ in} denotes the number of layer input units and \textbf{out} the number of layer output units. LReLU is the leaky rectified linear unit activation function with slope parameter $0.2$.}}{
		\scriptsize
		\centering
		\label{tab:nparams}
		\begin{tabularx}{.995\columnwidth}{|r  r  r r  r  r r| }
			\toprule
			\textbf{name} &\textbf{type} &\textbf{in} &\textbf{out}  &\textbf{input} &\textbf{activ}&\textbf{\#weight} \\ \midrule
			d1 & dense &  12 & 64 &$\mc I$&LReLU&832  \\ 
			d2 & dense &  64 & 128 &d1&LReLU &8320\\ 
			d3 & dense &  128 & 256 &d2&LReLU &33024 \\ 
			d4 & dense &  256 & 256 &d3&LReLU &65792 \\ 
			$\mc O$ & dense&  256 & 3 &d4&linear&771\\ \bottomrule
		\end{tabularx}
	}\hspace{1em}
		\ttabbox[.49\columnwidth]{\caption{Module architecture ${\nblc}$. \textbf{ in} denotes the number of input channels and \textbf{out} the number of output channels. Integer \textbf{k} denotes the size of the 1D-kernels and ReLU is the rectified linear unit activation function.}}{
			\scriptsize
			\centering
			\label{tab:nblc}
			\begin{tabularx}{.979\columnwidth}{|r  r r r r  r  r r| }
				\toprule
				\textbf{name} &\textbf{type} &\textbf{in} &\textbf{out} &\textbf{k}  &\textbf{input} &\textbf{activ}&\textbf{\#weight} \\ \midrule
				c1 & conv &  7 & 64&5 &$\mc I$&ReLU&2304  \\ 
				c2 & conv &  64 & 64&5 &c1&ReLU &20544\\ 
				c3 & conv &  64 & 128&5 &c2&ReLU &41088 \\ 
				c4 & conv &  128 & 128&5 &c3&ReLU &82048 \\ 
				c5 & conv &  128 & 256&5 &c4&ReLU &164096\\ 
				c6 & conv &  256 & 256&5 &c5&ReLU &327936 \\ 
				c7 & conv &  256 & 512& 5&c6&ReLU & 655872\\ 
				c8 & conv &  512 & 512&5 &c7&ReLU & 1311232\\ 
				$\mc O$ & conv&  512 & 1&5 &c8&linear&2561\\ \bottomrule
			\end{tabularx}
		}
	\end{floatrow}
\end{table}

\subsection{Loss Function Design}

\label{sec:loss}

In addition to the model architecture, the choice of the loss function has a major impact on model performance. In general, it is desirable to find a loss function that minimizes a task-dependent distance measure. Considering the preceding parameter prediction module, the mean absolute error (MAE) and the mean squared error are chosen to measure the distance between estimated and ground truth Sk, Vvv and Vmp values. To account for the different scales of these three parameter families, the ground truth parameters are standardized, i.e., each family is centered around its mean with unit standard deviation.
 In the subsequent module for BLC prediction, closer investigation is necessary to find an appropriate loss term.\\
 
Since the BLC (empirical quantile function) is the inverse of the empirical cumulative distribution function, comparing a predicted BLC to the corresponding ground truth is equivalent to comparing two probability distributions.  
A commonly considered metric for measuring the closeness of two probability distributions is the Wasserstein-1 distance \citep{villani2008,arjovsky2017,ramdas2017}, which is also called the earth mover's distance. The  Wasserstein-1 distance between two probability distributions $\dist_1$ and  $\dist_2$ is defined as $\wass_1(\dist_1, \dist_2) \triangleq \inf_{J\in\mathcal J(P_1,P_2)}\mathbb{E}_{(x,y)\sim J}\norm{x-y}$,
where the infimum is taken over the set of all joint probability distributions that have marginal distributions $\dist_1$ and $\dist_2$. The Wasserstein-1 distance can be interpreted  in the setting of optimal mass transport. In this setting, the aim is to find an optimal transfer plan to transport one mass distribution into another as cheaply as possible in reference to a given cost function \citep{adler2017}. 
It is important to note that, contrary to the standard  $L^p$-norms, the Wasserstein distance not only compares distribution  values point-wise  but also quantifies how far  the distributions have to be moved.

Wasserstein distances have several useful properties and dual representations, which makes iterative solution of the transport problem computationally feasible \citep{villani2008, arjovsky2017}. Moreover, given two one-dimensional probability distributions $\dist_1$ and $\dist_2$, the Wasserstein-1 distance simplifies to $\wass_1(\dist_1,\dist_2)  = \int_{0}^1| Q_1(z) -  Q_2(z)|dz$, where $ Q_1,  Q_2$ denote the corresponding quantile functions.  
Due to discretization of the BLCs, the Wasserstein-1 distance between ground truth depth distribution and predicted depth distribution corresponds to the component-wise $L^1$-distance  between the  corresponding discretized BLCs. This yields the component-wise MAE as the loss term for the model prediction $\hat \blc$ and the corresponding ground truth $\blc$:

\begin{linenomath*}
	\begin{equation} \label{eq:loss} 
	\lblc (\hat\blc, \blc) \triangleq 
	\frac{1}{K}\sum_{k=1}^K
	\bigl\lvert
	\hat\blc_k  -\blc_k
	\bigr\rvert.
 \end{equation}\end{linenomath*}

The final model for modality transfer $\mc T$ \eqref{eq:transfer} minimizes the empirical risk $\riskpar$ \eqref{eq:parrisk} in the first stage using the transformation $\Psi$ \eqref{eq:prepro}, the architecture  $\nparam$ described in Section~\ref{sec:paramestim} and training data $(\rgb_n, \blc_n)$ described in  Section~\ref{ssec:database}. Afterwards, the empirical risk $\riskblc$ \eqref{eq:blcrisk} is minimized using the Wasserstein-1 loss described in \eqref{eq:loss} and the architecture  $\nblc$ described in Section~\ref{sec:blcestim}.
A variety of efficient optimization algorithms exist for minimizing the risk functionals \eqref{eq:parrisk} and \eqref{eq:blcrisk}. For example, the optimization problems can be solved with a wide range of stochastic gradient descent implementations \citep{ruder2016,zeiler2012, kingma2014}.   In this work, both functionals are minimized using the Adam algorithm \cite{kingma2014}, an iterative minimization algorithm that permanently computes individual learning rates for all parameters and makes use of the current gradient direction for accelerated training.

\section{Results and Discussion}
\label{sec:results}

This section presents the results of BLC prediction from RGB reflection images recorded by a handheld device. According to Section \ref{ssec:database}, the custom-built database consists of a total of 422 registered modality pairs. 71 modality pairs are extracted from the database for evaluation while the remaining 351 pairs are used for training. Note that there is always a strict
separation between the liners, i.e., no liner contributes areas to the training data and the evaluation
data at the same time. This separation mimics use in a practical scenario where the model would also
be applied to liners that it has never seen during training. During splitting, it is ensured
that both sets have a similar liner operating hour distribution. The set of training data is augmented
to 1404 pairs by applying vertical flip and Gaussian blur to the reflection images while the corresponding BLCs remain unchanged.
The input  RGB reflection images cover a spatial area of \SI[]{1.9 x 1.9}{\mm} with a total of  \num{470 x 470} pixels (c.f. Section \ref{ssec:database}) and are resized to \num{512 x 512} pixels using linear interpolation. The corresponding BLC curve is calculated from a depth counterpart that exactly covers the RGB region and the amount of BLC sampling points $K$ is set to 512. 

\subsection{Parameter Selection and Training}

Parameter are fine-tuned by means of a five-fold cross-validation (5fCV). More specifically, data used for training is
split into five separate parts of nearly identical size. Again, sampling is carried out so that the areas from a single liner are not in different folds. During training, one part (validation data) is omitted
and then used for testing on unseen data while the model is optimized with the other four parts.
The process is repeated five times; each time another fifth is omitted. The parameter setup that achieves the best results on average over all validation sets becomes the final setup with which the model is subsequently trained on the entire training set. 

To measure the quality of the results achieved on the validation folds and the evaluation data, the similarity between a BLC prediction and its ground truth is assessed via the Wasserstein-1 distance \eqref{eq:loss}. The MAE is evaluated on the Sk, Vvv and Vmp values derived from the final BLC predictions and the corresponding ground truth curves. For Sk core roughness, the mean absolute percentage error (MAPE) is also calculated. Furthermore, the standard deviation of each error quantity is reported. Experiments showed that parameters extracted from the curves predicted by the second module are even more accurate than the parameters predicted in the first module.
Sk core roughness is measured on a \SI{}{\micro\metre} scale while volume parameters Vvv and Vmp are commonly specified in the unit  \SI[per-mode=symbol]{}{\milli\litre\per\metre\squared}. For the proposed training dataset in Section \ref{ssec:database}, the following statistic quantities (minimum $\mid$ 1st quartile $\mid$ median $\mid$ 3rd quartile $\mid$ maximum) are observed for these parameters:

\begin{itemize}
	\item \textbf{Sk} in \textbf{\SI{}{\micro \metre}}:\hspace{2.55em} 0.66 $\mid$ 0.95 $\mid$ 1.21 $\mid$ 1.42 $\mid$ 2.61
	\item \textbf{Vvv} in \textbf{\SI[per-mode=symbol]{}{\milli\litre\per\metre\squared}}:\hspace{0.4em} 0.05 $\mid$ 0.11 $\mid$ 0.15 $\mid$ 0.19 $\mid$ 0.39
	\item \textbf{Vmp} in \textbf{\SI[per-mode=symbol]{}{\milli\litre\per\metre\squared}}:
	 \num{1.2e-2} $\mid$ \num{1.95e-2} $\mid$ \num{2.58e-2} $\mid$ \num{3.04e-2} $\mid$ \num{6.97e-2}
\end{itemize}

Training of the preceding parameter prediction module $\nparam$ in \eqref{eq:parrisk} is performed within 30 epochs, where epochs denote how many times the total amount of training data was processed by the module for risk minimization. With the 5fCV the value \num{1e-3} turned out to be the best learning rate for the Adam optimizer in terms of training convergence. The LeakyReLU activation function with slope parameter 0.2 yielded the best prediction results in the 5fCV according to the MAE between Sk, Vvv and Vmp values. Normalization techniques after each dense layer do not improve the parameter prediction accuracy. Training stability is further enhanced through use of a learning rate scheduler: the actual learning rate is multiplied by a factor of $1/3$ if the validation loss does not decrease within five epochs.

In the second stage, the subsequent BLC prediction module $\nblc$ in \eqref{eq:blcrisk} that relies on the input data and the output of the first module is optimized. The loss function \eqref{eq:loss} is based on the Wasserstein-1 distance and minimized within 40 epochs. The best results in terms of 5fCV metrics were achieved with the ReLU activation function. Again, 5fCV indicated that a learning rate of \num{4e-4} in combination with the Adam optimizer and previously mentioned learning rate scheduler yields the best results. Another parameter for fine-tuning is the size of the one-dimensional convolution kernels; a kernel size of 5 yielded an appropriate trade-off between network performance and architecture efficiency. Insertion of an instance normalization step \citep{ulyanov2018} after each convolutional layer further improves validation accuracy.\\

\subsection{Quantitative Evaluation}

Table \ref{tab:cv} presents the 5fCV results for the joint BLC prediction framework using RGB data from a handheld device and a preceding parameter prediction module. Results of the unseen evaluation data (71 modality pairs) are displayed in the last row.
To avoid confusing scientific notation, the error quantities for the volume parameters Vvv and Vmp are scaled to \SI[per-mode=symbol]{}{\micro\litre\per\metre\squared} $\left(\times\ 10^3\right)$.
 Obviously, the validation metrics barely change across all five folds, which is an indication that the learning process is quite stable despite the different sets of training liners. The metrics on the evaluation set are quite similar to those of the 5fCV average metrics despite the lack of use of the evaluation liners during training and parameter fine-tuning. This suggests it is reasonable to apply the trained framework to unseen liners for reliable BLC prediction.
 
\begin{figure}[htb!!]
	\begin{floatrow}
		\ttabbox[0.99\columnwidth]{\caption{Results for the BLC prediction framework using RGB data from a handheld device and a preceding parameter prediction module. Reported metrics are the Wasserstein-1 distance and the MAE evaluated on the corresponding Sk, Vvv and Vmp values extracted from the BLC predictions (smaller is better). The standard deviation (std) of each error quantity is also given.}\label{tab:cv}
		}{

			\begin{tabularx}{.91\columnwidth}{l | c | c | c | c }
				\toprule
				\textbf{Run} & $\wass_1 \pm$ std&\textbf{Sk} MAE (MAPE) $\pm$ std&\textbf{Vvv} MAE $\pm$ std &\textbf{Vmp} MAE $\pm$ std \\ \midrule
				\multicolumn{5}{l}{Five-fold cross-validation}
				\\  \midrule
				1&$0.101\pm 0.060$&0.135 (\SI{11.1}{\percent}) $\pm$ 0.099 \SI{}{\micro\metre} 
				&$33.9\pm18.9$ \SI[per-mode=symbol]{}{\micro\litre\per\metre\squared}
				&$3.93\pm2.38$ \SI[per-mode=symbol]{}{\micro\litre\per\metre\squared}\\ \midrule
				
				2&$0.108\pm 0.058$ &0.171 (\SI{15.1}{\percent}) $\pm$ 0.109 \SI{}{\micro\metre}
				&$35.5\pm20.6$ \SI[per-mode=symbol]{}{\micro\litre\per\metre\squared} &$4.61\pm2.55$ \SI[per-mode=symbol]{}{\micro\litre\per\metre\squared}\\ \midrule
				
				3&$0.102\pm 0.055$&0.142 (\SI{11.3}{\percent}) $\pm$ 0.104 \SI{}{\micro\metre}
				&$33.7\pm19.5$ \SI[per-mode=symbol]{}{\micro\litre\per\metre\squared}
				&$4.24\pm2.55$ \SI[per-mode=symbol]{}{\micro\litre\per\metre\squared}\\ \midrule
				
				4&$0.102 \pm 0.077$&0.152 (\SI{10.9}{\percent}) $\pm$ 0.110 \SI{}{\micro\metre}
				&$35.5\pm22.9$ \SI[per-mode=symbol]{}{\micro\litre\per\metre\squared}
				&$4.08\pm2.90$ \SI[per-mode=symbol]{}{\micro\litre\per\metre\squared}\\ \midrule
				
				5&$0.106\pm 0.070$&0.163 (\SI{12.8}{\percent}) $\pm$ 0.125 \SI{}{\micro\metre}
				&$33.5\pm19.5$ \SI[per-mode=symbol]{}{\micro\litre\per\metre\squared}
				&$3.99\pm2.79$ \SI[per-mode=symbol]{}{\micro\litre\per\metre\squared}\\ \midrule

				\textbf{Avg.}&$\mathbf{0.104\pm0.064 }$&\textbf{0.153 (\SI{12.2}{\percent}) $\pm$ 0.109 \SI{}{\micro\metre}}&
				$\mathbf{34.4\pm20.3}$\textbf{ \SI[per-mode=symbol]{}{\micro\litre\per\metre\squared}}
				&$\mathbf{4.15\pm2.63}$\textbf{ \SI[per-mode=symbol]{}{\micro\litre\per\metre\squared}}
				\\ \midrule
				\multicolumn{5}{l}{Evaluation data}
				\\ \midrule
				1&$\mathbf{0.102\pm 0.070}$&\textbf{0.170 (\SI{13.5}{\percent}) $\pm$ 0.133 \SI{}{\micro\metre}}&
				$\mathbf{30.7\pm23.3}$\textbf{ \SI[per-mode=symbol]{}{\micro\litre\per\metre\squared}}
				&$\mathbf{4.15\pm2.81}$\textbf{ \SI[per-mode=symbol]{}{\micro\litre\per\metre\squared}}
				\\		\bottomrule
			\end{tabularx}
		}
\end{floatrow}
\end{figure}

\begin{figure}[htb!]
	\begin{floatrow}
		\ffigbox[.47\columnwidth]{
	\includegraphics[width=1.01\columnwidth]{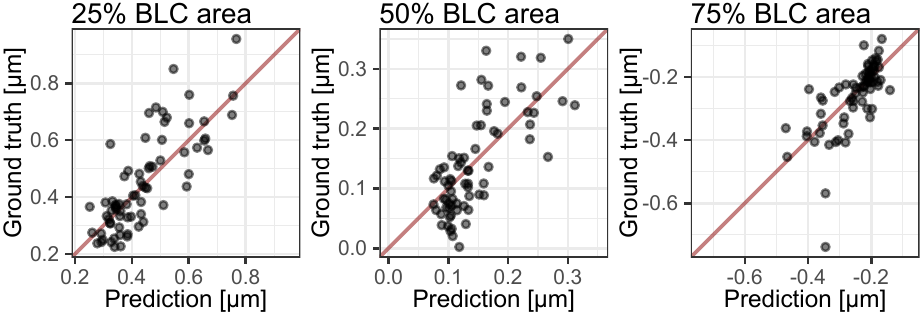}}{
	\caption{From left to right: The \SI{25}{\percent}, \SI{50}{\percent} and \SI{75}{\percent} BLC area quartiles are extracted from each BLC prediction of the evaluation set and plotted against the corresponding ground truth quartile values. In each case, perfect prediction of the quartiles corresponds to points on the diagonal.}
	\label{fig:quartiles}}\hspace{1em}
\ffigbox[.47\columnwidth]{
	\includegraphics[width=1.01\columnwidth]{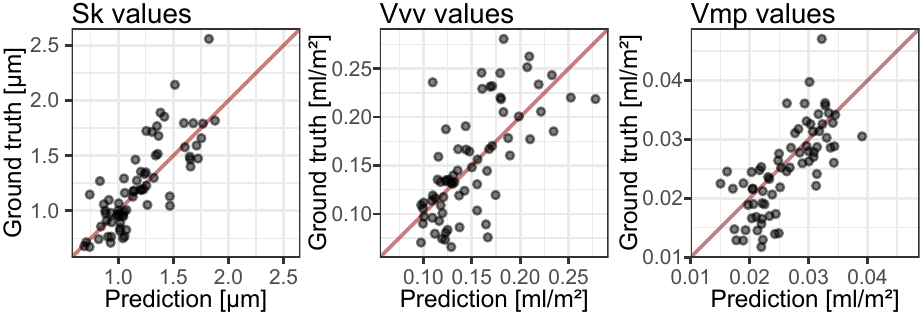}}{
	\caption{From left to right: The indicators Sk, Vvv and Vmp are extracted from each BLC prediction of the evaluation set and plotted against the corresponding indicators of the ground truth BLCs. In each case, perfect prediction of the indicators corresponds to points on the diagonal.}
	\label{fig:params}}
\end{floatrow}
\end{figure}

 Since the BLC provides valuable information on surface depth distribution, comparison of the BLC area quartile values is another good measure for performance of prediction. The values of the \SI{25}{\percent}, \SI{50}{\percent} and \SI{75}{\percent} BLC area quantiles are extracted from each BLC prediction in the evaluation set and plotted against their counterparts extracted from the corresponding ground truth BLCs (see Figure \ref{fig:quartiles}).
 A clear correlation is observed for all three quartiles, especially for the \SI{25}{\percent} and \SI{75}{\percent} values, which denote the transition
 from the core area to the peak and the valley region, respectively. However, the prediction model slightly overestimates lower ground truth values and underestimate higher ground truth values. Yet this may be advantageous, especially in the case where the evaluation ground truth data contains single outliers that the model should not overfit to. In sum, analysis of the quartile values further proves the reliability of the BLC predictions.


From an application-related point of view, subsequent extraction of BLC parameters is also a reasonable measure of prediction accuracy. As discussed in paragraph \ref{sec:paramestim}, the slope of the core area of the BLC is well identified by the Sk value, while valley and peak characteristics of the BLC are described by volume parameters Vvv and Vmp, respectively. 
Already considered in the preceding parameter prediction module, these variables are now used for further quantitative evaluation.
 In Figure \ref{fig:params}, the functional core roughness and the volume indicators extracted from BLC predictions of the unseen evaluation data are plotted
against their corresponding ground truths. Satisfactory correlations can be found for all three parameters; in particular, the calculation of the Sk and Vmp indicators from the BLC predictions provides quite accurate values in practical application.\\

\begin{figure}[htb!]
	\begin{floatrow}
		\ffigbox[.99\columnwidth]{
			\includegraphics[width=.9\columnwidth]{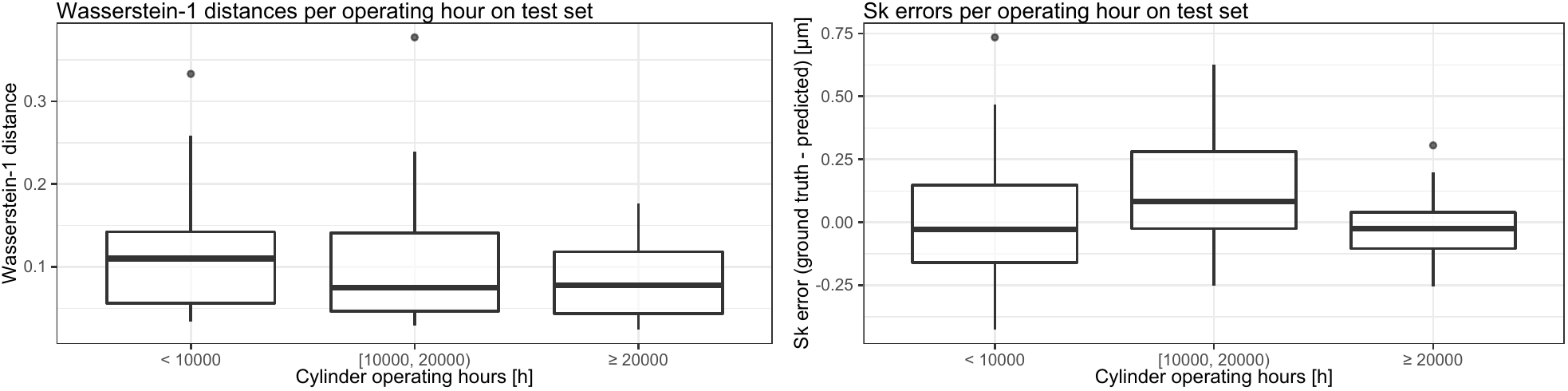}}{
			\caption{This boxplot diagrams visualize the dependency between the accumulated operating hours of a liner and the performance of the model on the unseen evaluation liner. The model performance is assessed via the Wasserstein-1 distance (left) and the Sk difference (right) between model predictions and corresponding ground truths.}
			\label{fig:oph}}\hspace{1em}
	\end{floatrow}
\end{figure}

In addition, the dependency between prediction accuracy and the amount of accumulated operating hours of the tested evaluation liners is investigated in Figure \ref{fig:oph}. The Wasserstein-1 distance between model predictions of the evaluation liners and corresponding ground truth BLCs decreases as the number of cylinder operating hours increases. It may be assumed that this is due to the decrease in BLC variance in heavily worn liners, which makes prediction easier for a model. In contrast, no trend can be identified for the residuals of the corresponding Sk parameters compared to the operating hours.

\section{Concluding Remarks}
\label{sec:conclusion}

Current engine liner evaluation protocols use sophisticated depth measurements that require disassembly and destruction of the engine components under investigation.  This research proposes the use of a simple surrogate RGB modality in combination with artificial neural networks to predict the BLC of a high-resolution depth measurement. It proposes a data-driven approach of joint fully-connected and convolutional network modules together with the Wasserstein-1 distance as a suitable loss function for BLC prediction. 
Training and comprehensive evaluation of the proposed model were carried out on a self-built database that consists of perfectly registered depth and RGB data acquired at great expense.
Quantitative evaluation of the predicted BLCs on unseen test data confirms both the plausibility and the reliability in practice when surrogate modalities are used in combination with advanced data-driven models.  
The proposed framework is also able to provide quite accurate estimates for core roughness (Sk), valley void volume (Vvv) and peak material volume (Vmp), which are very important in the assessment of surface oil retention capacity.
In addition, fine-tuning of parameters in a five-fold cross-validation process was required to stabilize model training and thus compensate for the small amount of available training data.  There are several directions in which to improve model performance in future research. First, more modality pairs will be acquired to further strengthen the generalizability and prediction accuracy of the proposed method. While the goal of this research to date has been to predict one-dimensional representations of the corresponding depth modality, the aim for the future is to generate the two-dimensional depth profile from the RGB image in both a supervised and an unsupervised manner. Finally, the parameter estimation module of the joint approach will be updated in order to extract even more information on different roughness and volume indicators out of the fast and inexpensive RGB modality, thus ensuring greater and more reliable support for subsequent prediction modules.

\section*{Funding}
The authors would like to acknowledge the financial support of the
"COMET - Competence Centres for Excellent Technologies” Programme
of the Austrian Federal Ministry for Climate Action, Environment,
Energy, Mobility, Innovation and Technology (BMK) and the Federal
Ministry for Digital and Economic Affairs (BMDW) and the Provinces of
Styria, Tyrol and Vienna for the COMET Centre (K1) LEC EvoLET. The
COMET Programme is managed by the Austrian Research Promotion
Agency (FFG) [grant number 865843].

\bibliographystyle{elsarticle-num}
%

\bibliography{DT04_publication1}

\end{document}